\documentclass[conference]{IEEEtran}
\IEEEoverridecommandlockouts
\usepackage{cite}
\usepackage{amsmath,amssymb,amsfonts}
\usepackage{algorithmic}
\usepackage{graphicx}
\usepackage{textcomp}
\usepackage{xcolor}
\usepackage{booktabs,tabularx}

\usepackage{tikz}
\usepackage{hyperref}

\newcommand\copyrighttext{%
  \footnotesize \textcopyright 2019 IEEE. Personal use of this material is permitted.
  Permission from IEEE must be obtained for all other uses, in any current or future
  media, including reprinting/republishing this material for advertising or promotional
  purposes, creating new collective works, for resale or redistribution to servers or
  lists, or reuse of any copyrighted component of this work in other works.\\
  Citation: Garcia-Ceja, E., Hugo, Å., Morin, B., Hansen, P. O., Martinsen, E., Lam, A. N., \& Haugen, Ø. (2019, November). Towards the Automation of a Chemical Sulphonation Process with Machine Learning. In 2019 7th International Conference on Control, Mechatronics and Automation (ICCMA) (pp. 352-357). IEEE.
  DOI: \href{https://doi.org/10.1109/ICCMA46720.2019.8988752}{10.1109/ICCMA46720.2019.8988752}}
\newcommand\copyrightnotice{%
\begin{tikzpicture}[remember picture,overlay]
\node[anchor=south,yshift=10pt] at (current page.south) {\fbox{\parbox{\dimexpr\textwidth-\fboxsep-\fboxrule\relax}{\copyrighttext}}};
\end{tikzpicture}%
}

\def\BibTeX{{\rm B\kern-.05em{\sc i\kern-.025em b}\kern-.08em
    T\kern-.1667em\lower.7ex\hbox{E}\kern-.125emX}}
\begin{document}

\title{Towards the Automation of a Chemical Sulphonation Process with Machine Learning\\
\thanks{Research leading to these results has received funding from the EU ECSEL Joint Undertaking under grant agreement no 737459 (project Productive4.0) and from the Research Council of Norway.}
}

\author{\IEEEauthorblockN{Enrique Garcia-Ceja, \AA smund Hugo, \\ Brice Morin}
\IEEEauthorblockA{\textit{Software and Service Innovation} \\
\textit{SINTEF Digital}\\
Oslo, Norway \\
\{enrique.garcia-ceja, \\ aasmund.hugo,brice.morin\}@sintef.no}
\and
\IEEEauthorblockN{Per Olav Hansen, Espen Martinsen}
\IEEEauthorblockA{\textit{}
\textit{Unger Fabrikker}\\
Fredrikstad, Norway \\
\{Per.Olav.Hansen,Espen.Martinsen\}@unger.no}
\and
\IEEEauthorblockN{An Ngoc Lam, \O ystein Haugen}
\IEEEauthorblockA{\textit{Department of Informatics} \\
\textit{\O stfold University College}\\
Halden, Norway \\
\{an.n.lam,oystein.haugen\}hiof.no}
}

\maketitle
\copyrightnotice

\begin{abstract}
Nowadays, the continuous improvement and automation of industrial processes has become a key factor in many fields, and in the chemical industry, it is no exception. This translates into a more efficient use of resources, reduced production time, output of higher quality and reduced waste. Given the complexity of today's industrial processes, it becomes infeasible to monitor and optimize them without the use of information technologies and analytics. In recent years, machine learning methods have been used to automate processes and provide decision support. All of this, based on analyzing large amounts of data generated in a continuous manner. In this paper, we present the results of applying machine learning methods during a chemical sulphonation process with the objective of automating the product quality analysis which currently is performed manually. We used data from process parameters to train different models including Random Forest, Neural Network and linear regression in order to predict product quality values. Our experiments showed that it is possible to predict those product quality values with good accuracy, thus, having the potential to reduce time. Specifically, the best results were obtained with Random Forest with a mean absolute error of $0.089$ and a correlation of $0.978$.
\end{abstract}


\section{INTRODUCTION}
\label{sec:introduction}

Enhancing chemical production processes can yield major economical and environmental benefits. A key measure of these enhancements is waste reduction. However, reducing waste proves challenging in flexible production processes that can be re-configured to accommodate different types of end products. While the process is being re-configured, there will typically be a transition with the production of an intermediate product, which does neither satisfy the requirements of the former, nor the new, product. In other words, these transition periods produce waste. Therefore, the ability to discover immediately when the output complies with the new product requirement, becomes pivotal to minimize waste from production.

The quality control during the aforementioned transition period used to be both fairly manual and conservative, involving manual sampling and observations. This impelled a transition period lasting undesirably long, with outputs satisfying the requirements of the new product going to waste. More and more, the chemical industry is investing into automated decision making processes, predominantly based on sensory data, so as to optimize their production and reduce waste~\cite{Ge2017}.
For example, an automated method to predict the quality of cobalt oxalate was reported by Zhang et al. \cite{ZHANG20131267}. The quality is based on the particle size which was estimated based on process variables such as reactor temperature, flow rate of ammonium oxalate, agitation speed, and so on.

In chemical production, measuring key process variables can be both difficult and expensive, due to complex non-linear relations and costly sensory equipment. Emerging from this, in combination with modern prediction modeling techniques, is the concept of \emph{soft sensing} \cite{Ge2017}. In soft sensing, the idea is to use easy-to-measure variables to predict the ones that are difficult to measure. Usually, the latter, are obtained by conducting offline lab analyses which are time consuming. Geng et al. proposed a new, more generalized soft sensor model, which they applied for accurately predicting the key variables of the Purified Terephthalic Acid (PTA) process \cite{GENG201738}. By developing and using an advanced neural network, they create a soft sensor model which is trained to predict the consumption of acetic acid on the basis of the PTA solvent system data.

Following this trend, Unger Fabrikker AS, a company producing chemicals used in active detergents, is currently investing in machine learning solutions to rationalize their production process, particularly with regards to waste reduction. Unger shifts between producing a variety of products during a normal week. This infers an approximately 30 minutes long transition period when shifting from a product to another, where parts of the chemical composition are unknown, and where Unger potentially produces waste. For Unger, it is critical to reduce this period of time.

To this extent, we trained different machine learning models in the quality control phase, with historical data gathered from chemical process parameters, in order to estimate the neutralization number (NT) which is a measure of the quality of the product. When applying a machine learning model with process parameters, to estimate this NT, we enter the realm of soft sensing and soft sensor models \cite{Ge2017}. Obtaining automated and accurate estimates of the product composition, and implicitly the quality, is vital towards the enhancement of the chemical sulphonation process. Similar to the work of Zhang et al. \cite{ZHANG20131267}, we predict the quality based on process variables. Based on our performance results, we see potential of improving the chemical process with the aid of soft sensor models.

The remainder of this paper is organized as follows. Section~\ref{sec:background} presents background information about machine learning and related work. Section~\ref{sec:process} presents an overview of the chemical process. Section~\ref{sec:data_collection} details the data collection procedure. In section~\ref{sec:experiments} we detail the conducted experiments and present the results. In section~\ref{sec:conclusions} we present our conclusions.

\section{Background}
\label{sec:background}

In this section we present an overview of machine learning and supervised learning. Then, we describe some related research works.

\subsection{Machine learning}

With the advent of information technologies, the amount of data that is generated everyday is growing at a fast pace. Trying to extract information and knowledge from that vast cumulus of data is a time consuming (if not impossible) task to do by hand. The computational power of machines has experienced a significant increase during recent years. That computational power can be used to analyze large quantities of data in an automatic manner. Machine learning methods and tools provide the means to automate the process of knowledge discovery and extraction from databases. \emph{Machine learning} can be thought of (but not limited to), as a set of computational algorithms that \emph{automatically} find interesting patterns and relationships from data. The key term here is: \emph{automatic}. Algorithms should be able to automatically scale without requiring explicit coded instructions. Kononenko \& Kukar define it as  ``The basic principle of machine learning is the automatic modeling of underlying processes that have generated the collected data."~\cite{kononenko2007machine}. Machine learning has two main types of models: \emph{supervised} and \emph{unsupervised}. Here, we will focus on \emph{supervised} methods, specifically, regression methods. More about unsupervised learning and other methods can be found in~\cite{witten2016data}. In \emph{supervised learning}, the algorithms are presented with a set of input variables and the corresponding output values from which they learn at training time. The aim is to find a mapping between input and output variables to generalize to unseen data points. When the output variable to be predicted is numeric it is called regression. When it is nominal it is called classification. In this work we used two supervised learning methods for regression: Random Forest and Neural Network.

A Random Forest~\cite{Breiman2001} is an ensemble model composed of several individual trees. In this case, regression trees. Each tree is built with different sub-samples of data and with random subsets of features at each tree split. The purpose of adding randomness is to generate de-correlated trees. The final prediction is obtained by averaging the output of all trees.

A Neural Network is a mathematical model that receives some input and produces an output. The traditional Neural Network architecture consists of a set of layers and units. Typically, there is an input layer, one or several hidden layers and an output layer. Each layer is composed of one or more units also known as neurons. As the name implies, the input layer receives the input values, it is the interface with the external world. Those input values are propagated through the hidden layers by applying several operations based on the weights between units. At the end, the output layer aggregates the outputs of the previous hidden layer and produces the final prediction. The parameters of the network (e.g., weights between units) are learned during training, usually with the gradient descent algorithm.

\subsection{Related work}

Soft sensors are predictive software models that make use of measured data from a given process, usually, industrial processes~\cite{KADLEC2009795}. A predictive model is a function that produces an output (the prediction) based on input variables. An example of a predictive model is a supervised learning algorithm like Random Forest. Soft sensors are mainly used to predict process variables that are related to process output quality~\cite{KADLEC2009795}. Those variables are typically estimated through manual off-line lab analyses which can be time consuming and/or expensive. Being able to estimate those variables more frequently, while reducing the required resources, is the main motivation of using soft sensors. Another use of soft sensors is as a back-up for physical sensors. If a given physical sensor fails, a soft sensor can take its place and start predicting estimated values while the physical sensor is fixed, thus, allowing the process to continue without major interruptions. Some of the advantages of soft sensors are: they are a low cost alternative compared to expensive hardware, they can work in parallel with physical sensors, they can provide real time estimations, etc.~\cite{fortuna2007soft}.

A natural application of soft sensing technologies is within the chemical industry since online monitoring~\cite{ge2013review} and waste reduction are key elements during the process. In the previously mentioned work of Zhang et al.~\cite{ZHANG20131267}, the authors implemented an online quality prediction system for a cobalt oxalate synthesis process. The final quality of products such as cutting tools and batteries, depend on the size and morphology of cobalt powders, thus, being able to measure particle size becomes important. Average particle size is measured by means of an offline analysis which is usually conducted one time per day. In order to reduce time, the authors proposed a soft sensor method based on least squares support vector regression that achieved a root mean squared error of $0.052$.

One of the problems with predictive models is that they need several representative data points during the training phase. A data point is composed of the input variables and the expected output (label). Usually, input variables are easy to obtain but the output variables (labels) require more effort, e.g., conducting an offline analysis. Because of this, many databases contain huge amounts of data points with only input variable values but empty labels and just a small proportion of data points contain both (input variables and labels). To address this, Bao et al.~\cite{bao_co-training_2015} proposed a method based on  co-training and partial least squares. In machine learning, co-training is a method that uses both, labeled and unlabeled data points to train a model~\cite{blum_combining_1998}. The authors tried their method on the Tennessee Eastman process benchmark to predict purge gas stream based on easy-to-measure variables. Their method presented significant improvements compared with the traditional method when labeled data was readily available. 
Another research work where easy-to-measure variables are used to predict hard-to-measure variables is presented in \cite{de2016soft}. Here, the authors used a neural network to predict primary chemical oxygen demand, nitrogen content and total suspended solids at a waste-water treatment plant. In this work, we follow a similar approach to predict the NT value which is a measure that represents the quality of the product. The predictive models are trained based on process variables measured during the chemical process. The production chemical process and related variables are presented in the next section.

\section{Production chemical process}
\label{sec:process}

\begin{figure*}[ht!]
\centering
\includegraphics[width=1.0\textwidth]{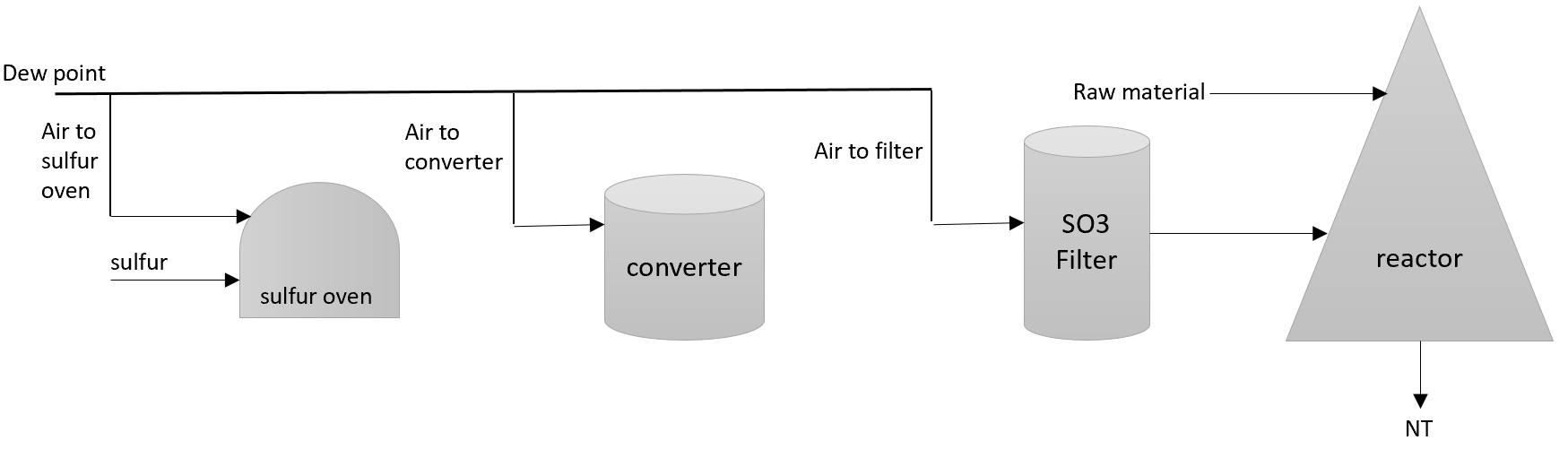}
\caption{Sulphonation process.}
\label{fig:process}
\end{figure*}

The chemical production process at Unger Fabrikker AS is a sulphonation process used for the manufacture of active detergents. A sulphonation reaction is based on different sulphonation reagents~\cite{ref1}, and the process at Unger is based on Sulphur burning and conversion of SO\textsubscript{2} gas to SO\textsubscript{3} gas. The SO\textsubscript{3} gas is diluted with air and mixed with organic liquid (raw material) in a liquid-gas reactor. The dew point of the air is a crucial part of the Sulphur burning and the conversion of SO\textsubscript{2} gas to SO\textsubscript{3} gas. The dew point should be at least $-60$ C to prevent the formation of sulphuric acid mist. The output from the reactor is a sulphonic acid with a variety of qualities based on the type of organic liquid used in the sulphonation process. The whole process is depicted in Figure~\ref{fig:process}.

The NT-value measures the reaction quality i.e, how much of the organic liquid is sulphonated, which is determined by the neutralization number (NT), and defines how many mg KOH (Kalium Hydroxid) are needed to neutralize one gram of sulphonic acid~\cite{ref2}. To define the neutralization number, the titration method by Karl Fischer is used~\cite{ref3}.
Unger Fabrikker has several transitions between different products during one week and therefore the neutralization number will differ in respect of which product they are producing. To check the performance of the transition and the quality in producing time there is a need for analyses of the product. The analyses results will have a delay of approximately thirty minutes. This means that the production will be in a   ``blind spot'' (historical data based on experience) during this waiting time for analyses results. However, to have a continuous measurement, the operator will have a confidence ability of analyses results by using machine learning to predict the neutralization number. Further, the number of analyses taken in the local laboratory could be reduced to more than a half.

\section{Data collection}
\label{sec:data_collection}

\begin{figure}[ht!]
\centering
\includegraphics[width=1.0\columnwidth]{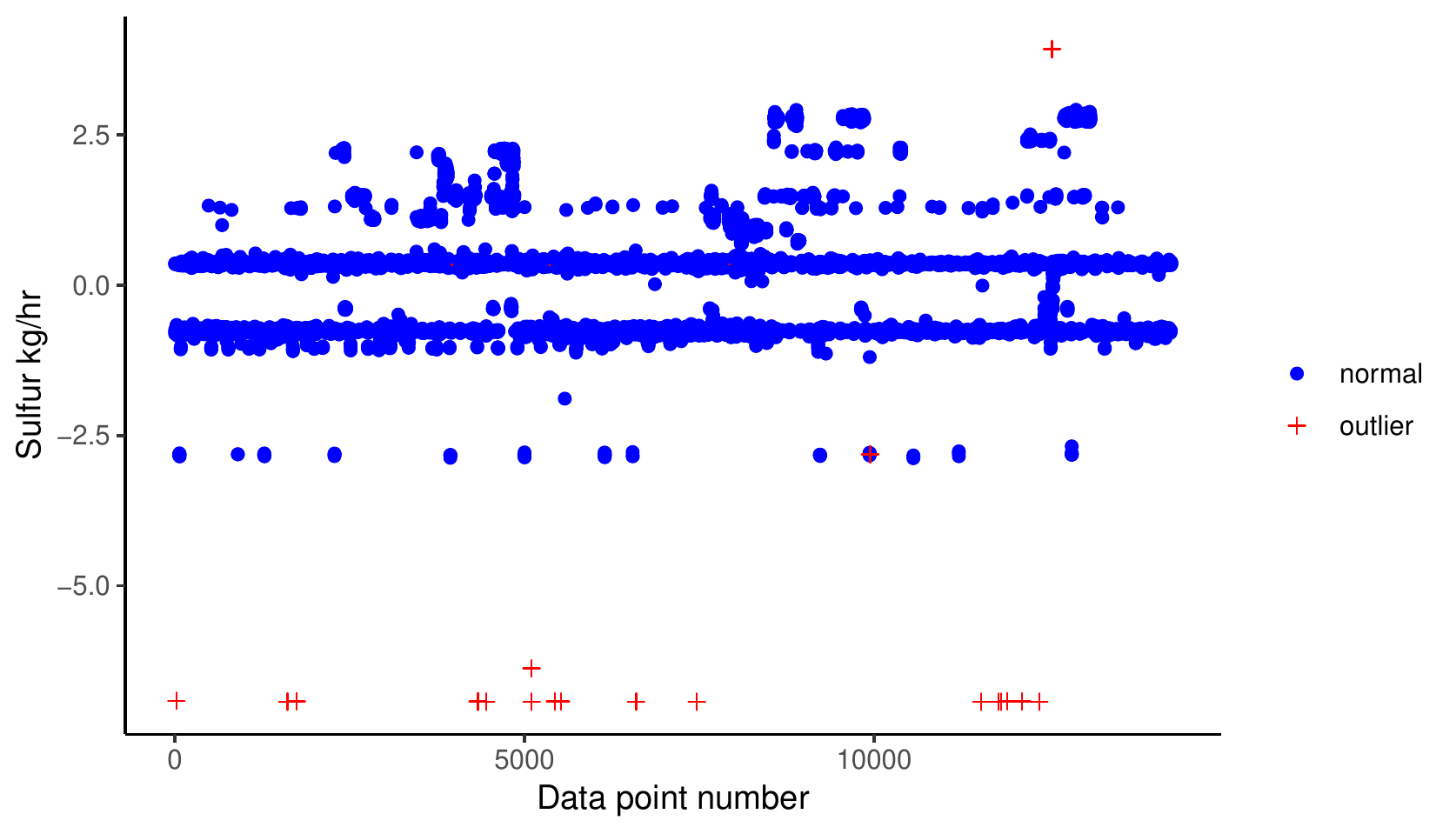}
\caption{Outliers plot with respect to the \emph{Sulfur} parameter. The x axis is the data point number with no particular order.}
\label{fig:outliers}
\end{figure}

\begin{figure}[ht!]
\centering
\includegraphics[width=1.0\columnwidth]{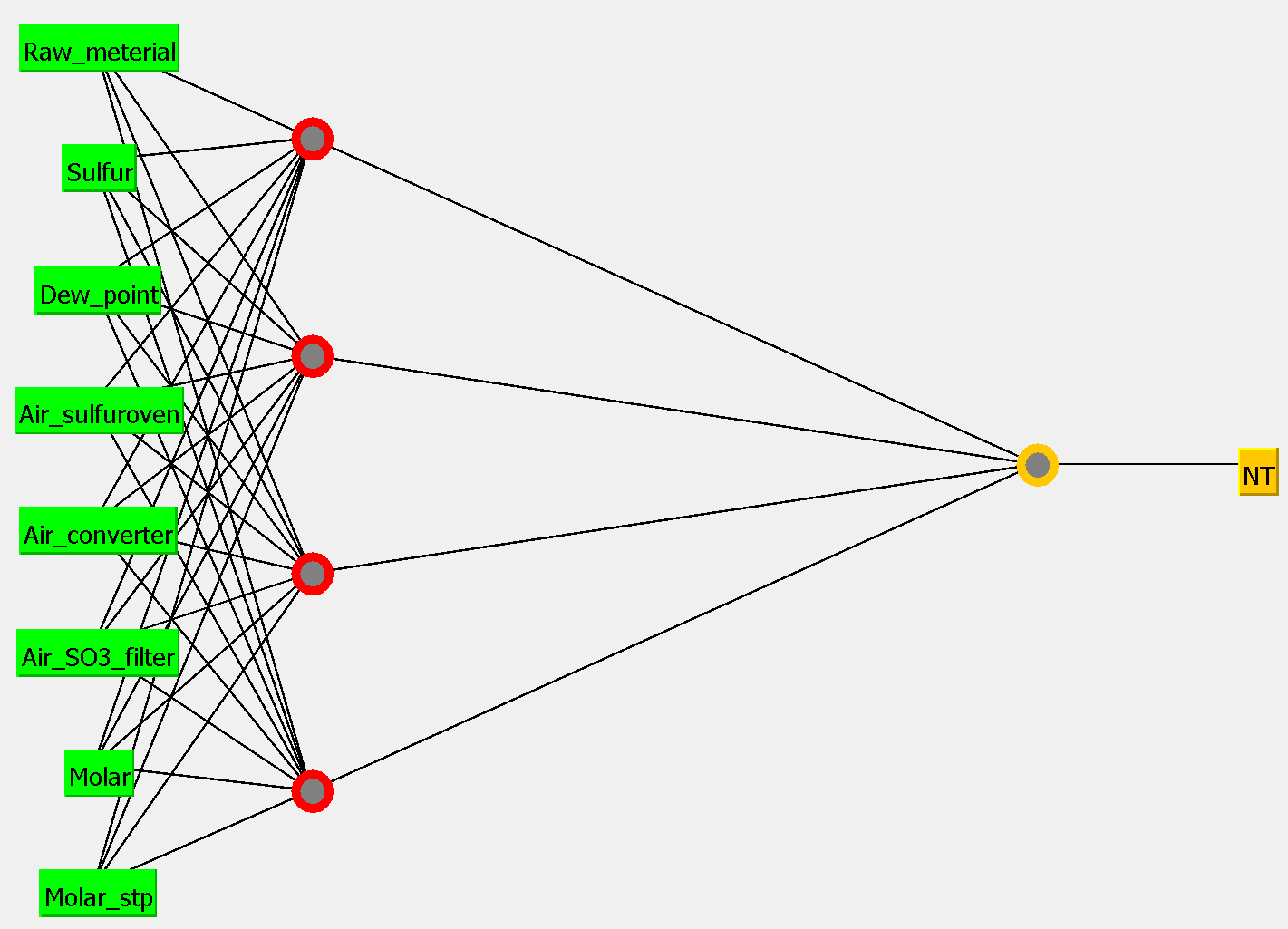}
\caption{Architecture of the used neural network. It consists of 1 input layer, 1 hidden layer with 4 units and an output layer with 1 unit.}
\label{fig:nn}
\end{figure}

It is important that the quality of the products are within the limits regarding the analysis result. To fulfill this purpose, the operator takes samples from the production line and analyze them with the help of the titration method. After the results are taking place, the operator  adjusts the parameters in the production to reach the quality specification. The time-line between the taken sample and the result are approximately 30 minutes. The analysis results are stored in a database with a time-stamp.
All of the process parameters are stored into a historian database so that these values can be inspected later for analysis. Process parameters will be set as points to regulators and new values from the production. Typical process values will be temperatures, pressures, flow and potential of hydrogen (pH). In this case, 8 process parameters were used to predict the NT value:

\begin{enumerate}
\item \emph{Raw-material.} This is the quantity of organic material in kg/hr.
\item \emph{Sulfur.} This is the amount of sulfur in kg/hr.
\item \emph{Dew-point.} This is the value of how dry the air is, measured in temperature.
\item \emph{Air-sulfur-oven.} This is the quantity of air injected into the sulfur oven nm\textsuperscript{3}/hr.
\item \emph{Air-converter.} This is the amount of air injected into the converter in nm\textsuperscript{3}/hr.
\item \emph{Air-SO3-filter.} This is the quantity of air injected into the SO3 filter in nm\textsuperscript{3}/hr.
\item \emph{Molar.} This is the mol rate.
\item \emph{Molar-stp.} This is the molar weight.
\end{enumerate}

In order to preserve data confidentiality, the variables were normalized by subtracting the mean and dividing by the standard deviation from each of the data points. In total, the dataset contains $14,252$ data points. From those, there are $23$ outliers, i.e, analyses with anomalous values in one or more parameters. Because of those outliers, two datasets were created. One with outliers and another one removing the outliers. Currently, the outliers are manually identified by an experienced engineer. For future work, we will explore methods to automatically detect those outliers. Figure~\ref{fig:outliers} shows a plot of the \emph{Sulfur} parameter with outliers marked with the '+' symbol. The x axis represents the data point number with no particular ordering. The y axis represents the amount of sulfur in kg/hr.

\section{Experiments and results}
\label{sec:experiments}

For the experimental phase, we considered two settings: 1) dataset with outliers, and 2) dataset without outliers. For each setting, we trained 3 different machine learning models: Random forest \cite{Breiman2001}, linear regression and a neural network.
To train the linear regression model, we used the \emph{lm} function which is part of the base R programming language. For the random forest model, we used the \emph{randomForest} R library~\cite{randomForestLib}. The random forest consists of 100 trees. The neural network architecture consists of an input layer of size $8$ which corresponds to the $8$ process parameters. Then, it has a hidden layer of $4$ sigmoid units and an output layer of a single linear unit that produces the final prediction of the NT value. We used the WEKA software~\cite{weka} to train the neural network with the default learning rate of $0.3$ and a batch size of $100$. Figure~\ref{fig:nn} shows a graphical representation of the network's architecture.
The models use the eight parameters described in section~\ref{sec:process} as input features. Based on those input features, the models predict the NT number. As a baseline, we also used a \emph{dummy} model that always predicts the mean NT value from the training set regardless of the input data.

The data points were randomly split into two subsets: 70\% for the train set and the remaining 30\% was used as the test set.  Tables~\ref{tab:results_out}-\ref{tab:results_noout} show the prediction results on the test set for the two settings: with and without outliers, respectively. The tables show the mean absolute error (MAE), root mean squared error (RMSE) and the Pearson correlation (Correlation).

\begin{table}[ht!]
  \centering
  \caption{Results on dataset with outliers.}
    \begin{tabular}{cccc}
    \toprule
          \textbf{} & 
          \textbf{MAE} &
          \textbf{RMSE} &
          \textbf{Correlation} \\
    \midrule
    \textbf{Random Forest} & 
    \multicolumn{1}{c}{\underline{0.091}} & \multicolumn{1}{c}{\underline{0.295}} & \multicolumn{1}{c}{\underline{0.955}} \\
	\midrule
	\textbf{Neural Network} & 
    \multicolumn{1}{c}{0.232} & \multicolumn{1}{c}{0.399} & \multicolumn{1}{c}{0.940} \\
    \midrule
    \textbf{Linear regression} & 
    \multicolumn{1}{c}{0.146} & \multicolumn{1}{c}{0.360} & \multicolumn{1}{c}{0.932} \\
    \midrule
    \textbf{Mean value} & 
    \multicolumn{1}{c}{0.700} & \multicolumn{1}{c}{0.999} & \multicolumn{1}{c}{0.0} \\
    \bottomrule
    \end{tabular}
  \label{tab:results_out}
\end{table}

\begin{table}[ht!]
  \centering
  \caption{Results on dataset without outliers.}
    \begin{tabular}{cccc}
    \toprule
          \textbf{} & 
          \textbf{MAE} &
          \textbf{RMSE} &
          \textbf{Correlation} \\
    \midrule
    \textbf{Random Forest} & 
    \multicolumn{1}{c}{\underline{0.089}} & \multicolumn{1}{c}{\underline{0.205}} & \multicolumn{1}{c}{\underline{0.978}} \\
	\midrule
	\textbf{Neural Network} & 
    \multicolumn{1}{c}{0.115} & \multicolumn{1}{c}{0.236} & \multicolumn{1}{c}{0.972} \\
    \midrule
    \textbf{Linear regression} & 
    \multicolumn{1}{c}{0.118} & \multicolumn{1}{c}{0.245} & \multicolumn{1}{c}{0.969} \\
    \midrule
    \textbf{Mean value} & 
    \multicolumn{1}{c}{0.704} & \multicolumn{1}{c}{0.999} & \multicolumn{1}{c}{0.0} \\
    \bottomrule
    \end{tabular}
  \label{tab:results_noout}
\end{table}

\begin{figure}[h!]
\centering
\includegraphics[width=1.0\columnwidth]{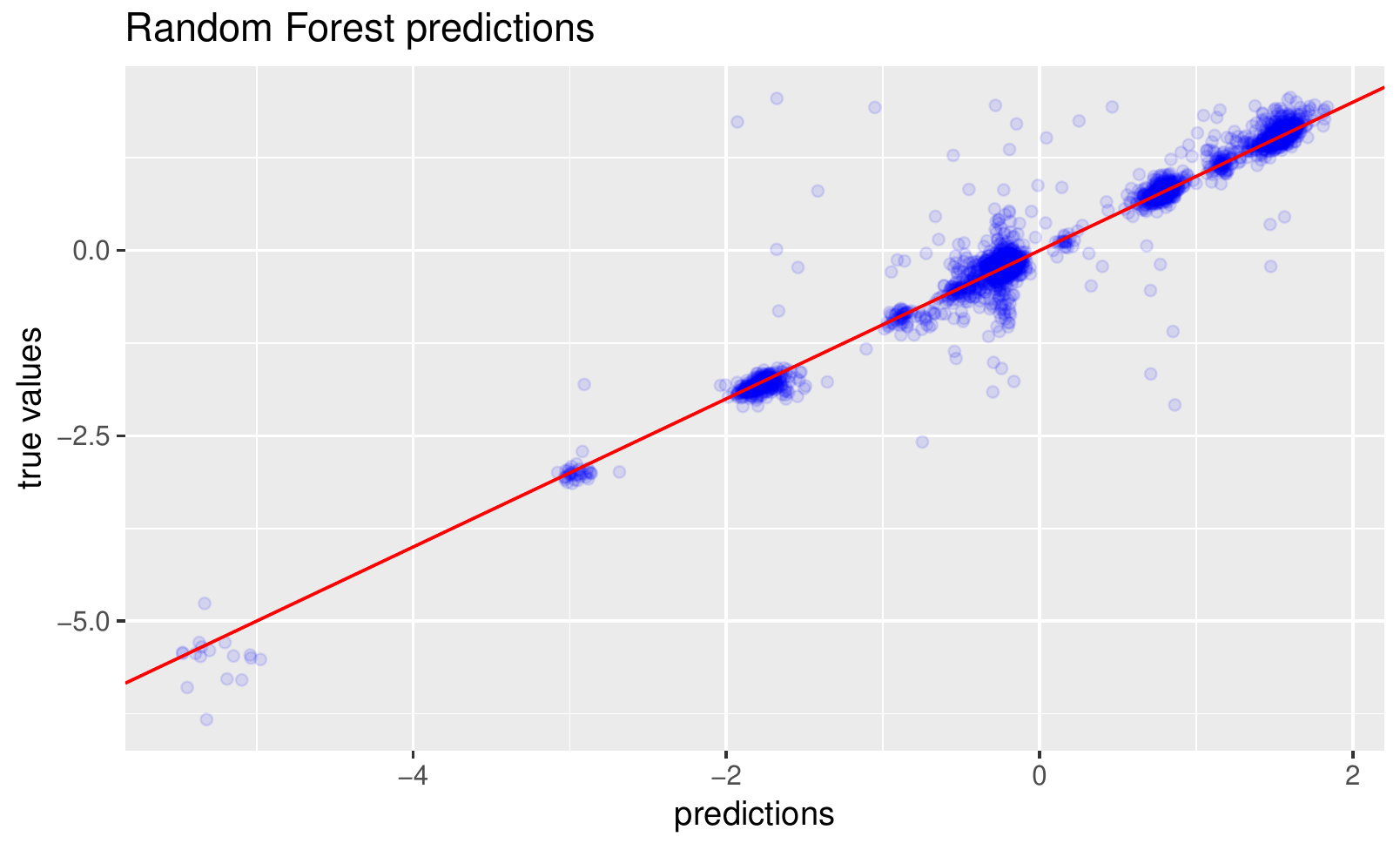}
\caption{Predictions with random forest.}
\label{fig:predictions}
\end{figure}

From these tables it can be seen that Random Forest produced the best results in terms of MAE, RMSE and correlation. It can also be noted that the results were better when removing the outliers. The minimum MAE (0.089) and RMSE (0.205) were achieved by Random Forest. The maximum correlation of 0.978 was also achieved by Random Forest. All algorithms outperformed the baseline mean value. Figure~\ref{fig:predictions} shows the predicted NT values versus the true NT values when using Random Forest.

\subsection{Variable importance analysis}

\begin{figure}[h!]
\centering
\includegraphics[width=1.0\columnwidth]{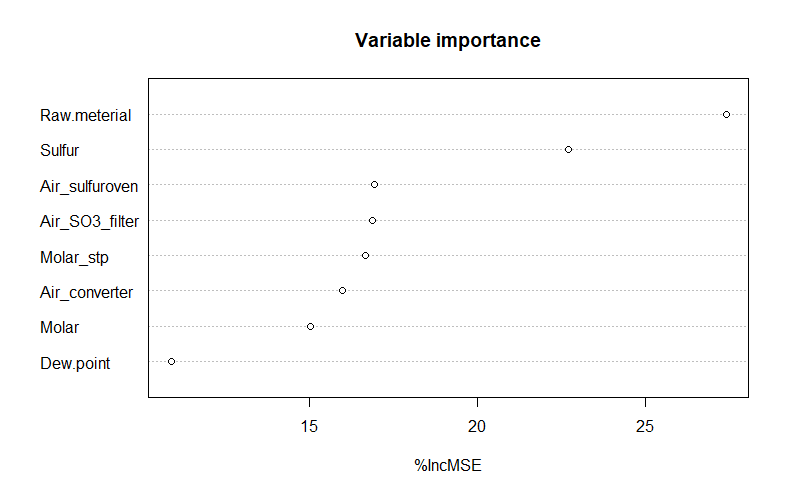}
\caption{Variable importance plot produced by random forest. Top variables are the most important based on the mean squared error. In this case, Raw-material and Sulfur. The x axis is the percent increase of the mean squared error (MSE).}
\label{fig:importance}
\end{figure}

Given that Random Forest achieved the best results, we further investigated which were the most important variables when predicting the NT value with such a model. Figure~\ref{fig:importance} shows the variable importance plot generated by Random Forest. The variables at the top are the most important ones based on the increase of the mean standard error. The x axis shows the percent increase of the mean standard error which is calculated based on the out-of-bag data. For each tree, the prediction error on the out-of-bag portion of the data is computed. Then, the same is done after permuting each predictor variable. The difference between the two are averaged over all trees and normalized by the standard deviation of the differences. From this plot, we can see that \emph{Raw\_material} and \emph{Sulfur} are the most important variables based on this criterion.

It is also important to understand what is the relationship between the predictors and the output variable of a model. Since Random Forest is an ensemble method composed of several decision trees, it would be a difficult task to perform such an analysis with each individual tree and then aggregating the results. Fortunately, there exist model agnostic methods that can help to understand these variables' relationships \cite{molnar2018,friedman2001greedy}. In this case we used partial dependence plots (PDPs) which can be used to visualize the marginal effect of a set of predictors (usually, one or two) on the predicted outcome (NT in this case). The partial dependence of a set of features of interest $z_s$ can be estimated by

\begin{equation} \label{eq:moving_average}
{f_s}(z_s) = \frac{1}{n}\sum\limits_{i = 1}^{n} {\hat{f}(z_s,z_{i,c})},
\end{equation}

where $\hat{f}(x)$ is the prediction function and $z_c$ is the compliment of $z_s$. Thus, $z_{i,c} (i=1,\dots ,n)$ are the values of $z_c$ that occur in the dataset~\cite{friedman2001greedy}.

\begin{figure}[h!]
\centering
\includegraphics[width=1.0\columnwidth]{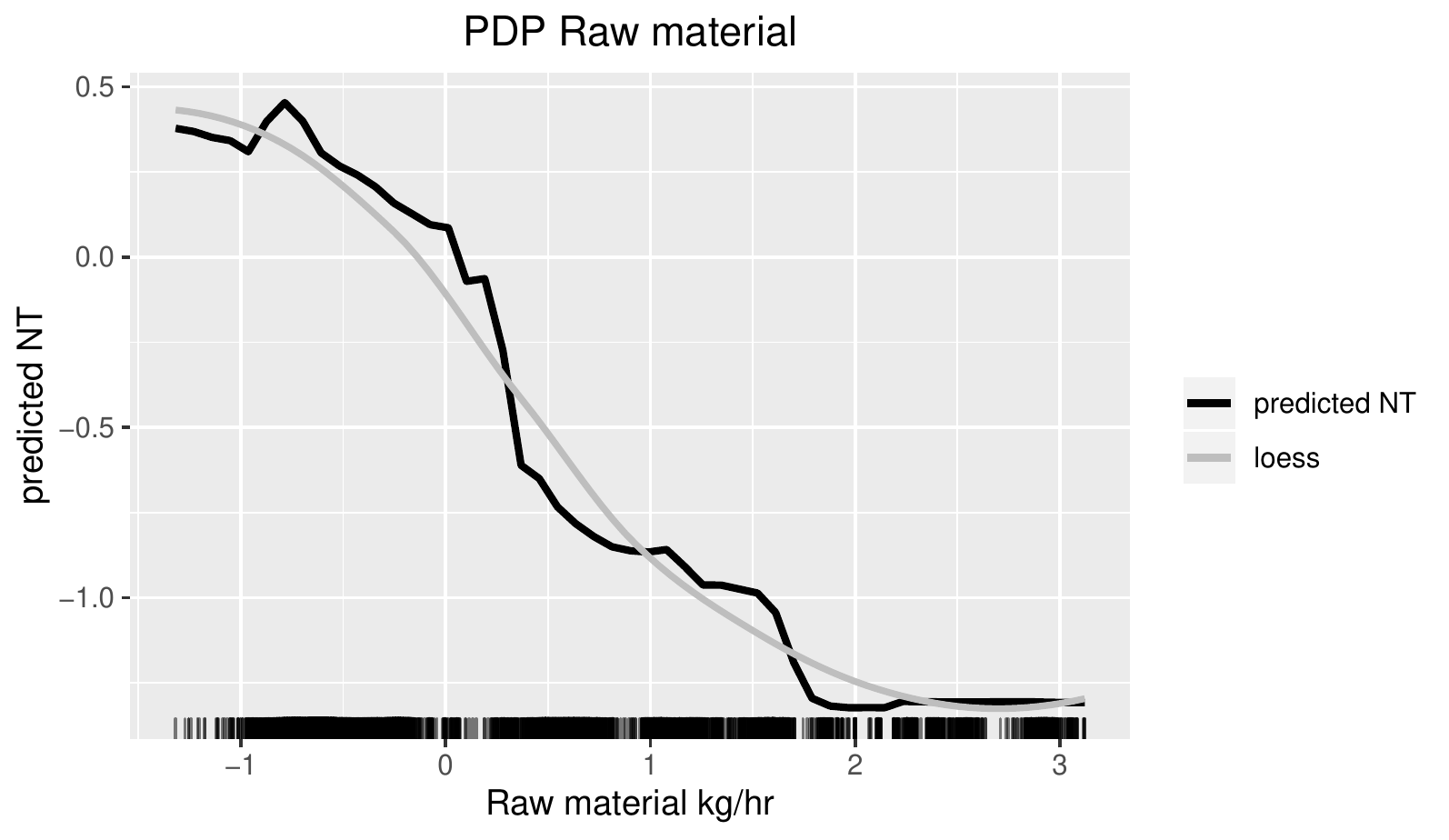}
\caption{Partial dependence plot with random forest. The black line represents the actual predictions whereas the gray line is a smoothed version to ease the visualization of the overall data trend. All values are shown after normalization.}
\label{fig:pdp_rastoff}
\end{figure}

Figure~\ref{fig:pdp_rastoff} shows the PDP for the \emph{Raw\_material} variable. The \emph{x} axis is the \emph{Raw\_material} value and the \emph{y} axis is the predicted NT value by Random Forest. Both variables' values are after normalization. Here we can see that as the \emph{Raw\_material} increases, the predicted NT value decreases. This negative relationship was validated by computing the pearson correlation coefficient which was $-0.4$. The black line represents the random forest's predictions when varying the value of raw material but keeping all other variables fixed. The gray line is a smoothed version of the original predictions that helps to visualize the trend.

\section{Conclusions}
\label{sec:conclusions}

In this work we explored the use of machine learning models to predict the quality of the product from chemical process parameter values. We trained and evaluated different models including Random Forest, Neural Network and linear regression. We did this in two different settings. One with outliers and the second one removing outliers from the dataset. The best results were obtained with Random Forest on the dataset without outliers with a positive correlation of $0.978$. Based on our preliminary results, we see that there is potential for implementing a system capable of automating the process by reducing manual efforts and analyses' time. In this case, all algorithms performed better without outliers. These results suggest that one should devote some effort in reducing noisy data points. In the present work, outlier removal was made manually. For future work, we will explore automatic methods to detect and remove outliers. Furthermore, we will also explore the use of deep learning models since in recent years they have proven to produce state of the art results in different domains.



\bibliographystyle{IEEEtran}
\bibliography{IEEEabrv,references}

\end{document}